\documentclass[runningheads]{llncs}
\usepackage{graphicx}

\usepackage{amsmath}
\usepackage{amssymb}
\usepackage[utf8]{inputenc}
\usepackage{xcolor}
\usepackage{algorithm} 
\usepackage{algpseudocode} 
\usepackage{subcaption}
\usepackage{setspace}
\usepackage{cuted}
\usepackage{hyperref}
\usepackage{bbm}

\begin{document}
%
%
\title{Case Studies for Computing Density of Reachable States for Safe Autonomous Motion Planning}

\titlerunning{Computing Density of Reachable States for Safe Motion Planning}
%
\author{Yue Meng\inst{1} \and
Zeng Qiu\inst{2} \and
Md Tawhid Bin Waez\inst{2} \and
Chuchu Fan\inst{1}}
%
%
\institute{Massachusetts Institute of Technology, USA \and
Ford Motor Company, USA}

\maketitle              

\begin{abstract}
Density of the reachable states can help understand the risk of safety-critical systems, especially in situations when worst-case reachability is too conservative. Recent work provides a data-driven approach to compute the density distribution of autonomous systems' forward reachable states online. In this paper, we study the use of such approach in combination with model predictive control for verifiable safe path planning under uncertainties. We first use the learned density distribution to compute the risk of collision online. If such risk exceeds the acceptable threshold, our method will plan for a new path around the previous trajectory, with the risk of collision below the threshold. Our method is well-suited to handle systems with uncertainties and complicated dynamics as our data-driven approach does not need an analytical form of the systems' dynamics and can estimate forward state density with an arbitrary initial distribution of uncertainties. We design two challenging scenarios (autonomous driving and hovercraft control) for safe motion planning in environments with obstacles under system uncertainties. We first show that our density estimation approach can reach a similar accuracy as the Monte-Carlo-based method while using only 0.01X training samples. By leveraging the estimated risk, our algorithm achieves the highest success rate in goal reaching when enforcing the safety rate above 0.99.
\end{abstract}

\section{Introduction}

Verifying and enforcing the safety of the controlled systems is crucial for applications such as air collision avoidance systems~\cite{julian2021reachability}, space exploration~\cite{kornfeld2014verification}, and autonomous vehicles. It is still a challenging problem to perform online verification and controller synthesis for high-dimensional autonomous systems involving complicated dynamics and uncertainties because of the scalability issue in verification and the absence of the analytical form to describe system trajectories. 

Reachability analysis is one of the main techniques used for rigorously validating the system's safeness~\cite{ivanov2019verisig,tran2020nnv,fan2020reachnn,hu2020reach,everett2021efficient} and controller synthesis~\cite{shkolnik2009reachability,gerdts2011avoidance,kousik2020bridging,manzinger2020using}. In reachability analysis, one computes the reachable set, defined as the set of states where the system (with the control inputs) can be driven to from the initial conditions, under the system dynamics and physical constraints. Take the aircraft collision avoidance system as an example: the system safety can be guaranteed if all the future space that the airplane can reach (under physical constraints) will not overlap with obstacles. However, computing the reachable states is proved to be undecidable in general (e.g., polynomial dynamical systems with degrees larger than 2)~\cite{hainry2009decidability} and is also empirically time-consuming, limiting applications to simple dynamics (e.g., linear systems) or low-dimension systems.


Besides, using worst-case reachability for safety analysis will usually return a binary result (``yes" or ``no"), regardless of the initial state distribution and the uncertainty in the systems. The focus on the ``worst-case" makes the corresponding reachability-based planning methods ``conservative" or ``infeasible" when the initial state has a large uncertainty. Consider a robot navigating in an environment with obstacles and state uncertainty - when a collision is inevitable in the worst case (though the worst case is a rare event), the planning algorithm will fail to return any safety-guaranteed control policies but to let the robot stop. Hence in those cases, we need a way to quantify the risk/probability of the undesired event (e.g., collision) happening and guide controller designs. 



\begin{figure}[!htbp]
\centering
\begin{subfigure}[b]{0.48\textwidth}  
\includegraphics[width=1.0\textwidth]{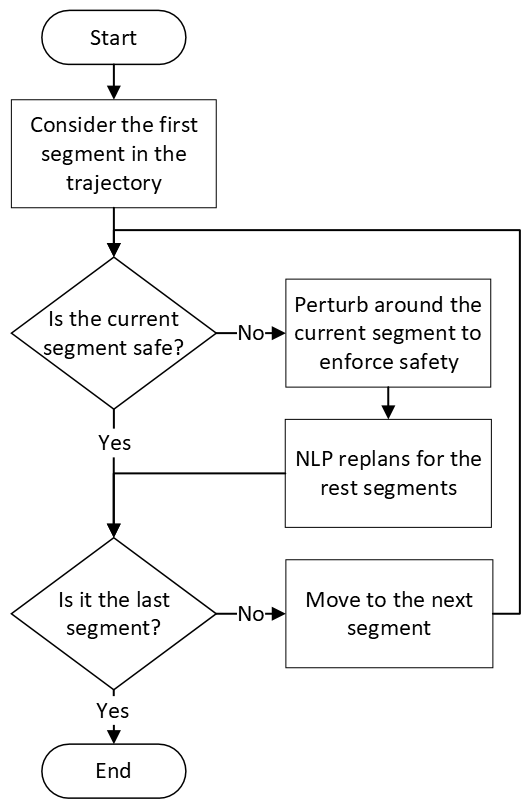} \hfill
\caption{Flow chart of the algorithm}
\end{subfigure}
\begin{subfigure}[b]{0.28\textwidth}  
\includegraphics[width=1.0\textwidth]{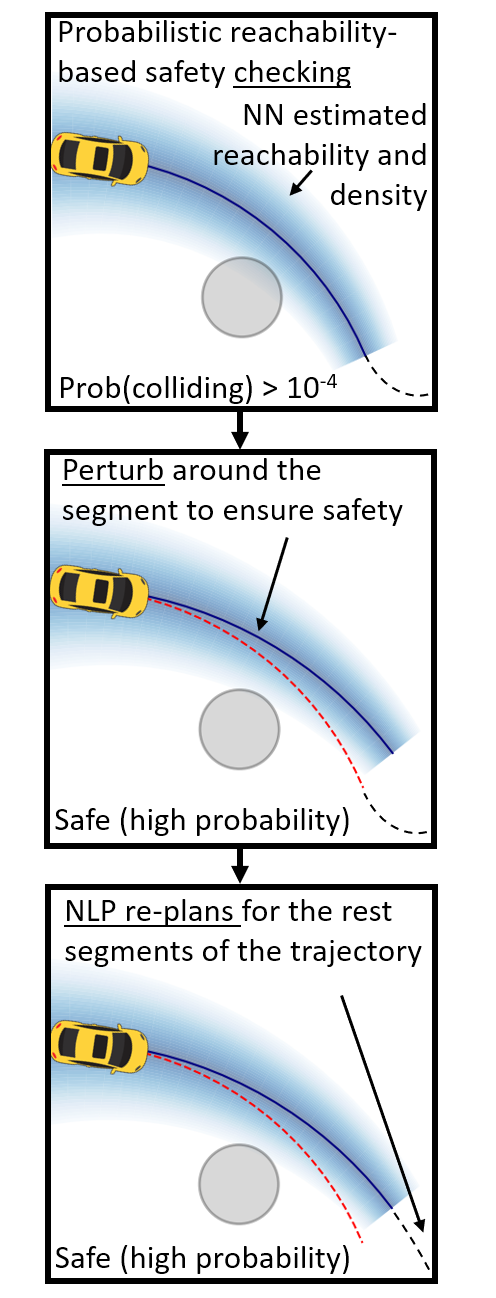} \hfill
\caption{Planning process}
\end{subfigure}\\
\caption{The safe planning algorithm.}
\label{fig:diagram}
\end{figure}

In this paper, we present a probabilistic and reachability-based planning framework for safety-critical applications. Inspired by~\cite{meng2021learning}, we first learn the system flow maps and the density evolution by solving the Liouville partial differential equation (PDE) using Neural Networks from collected trajectory data. Instead of using the exact reachability analysis tool~\cite{vincent2020reachable} for reachable states probability estimation, we use Barycentric interpolation~\cite{hormann2014barycentric}, which can handle more complicated systems (dimension $>4$) and sharply reduces the processing time compared to~\cite{meng2021learning}. In addition, by picking different numbers of sampled points, our algorithm can flexibly control the trade-off between estimation efficiency and accuracy. Leveraging this density estimation technique, our planning framework (illustrated in Fig.~\ref{fig:diagram}) verifies the safety of the system trajectory via a segment-by-segment checking. If one segment becomes unsafe, we perturb around the reference trajectory to find a safe alternative, and plan for the rest of the trajectory. The process repeats until all segments are enforced to be safe. 

We conduct experiments on two challenging scenarios (autonomous car and hovercraft control with uncertainties). Our estimated reachable states density distribution is informative as it reflects the contraction behavior of the controllers and highlights the places that the system is more likely to reach. Quantitatively, compared to Monte Carlo density estimation, our approach can achieve a similar accuracy while only using 0.01X training samples. We test our density-based planning algorithm in 20 randomly generated testing environments (for each system), where we achieve the highest success rate in goal reaching with high safety rate (measured by one minus the collision rate) compared to other baselines.

Our contributions are: (1) we are the first group to study the use of learned reachability density in safe motion planning to ensure probabilistic safety for complicated autonomous systems, (2) our approach can estimate state density and conduct safe planning for systems with nonlinear dynamics, state uncertainty, and disturbances, and (3) we design both qualitative and quantitative experiments for two challenging systems to validate our algorithm being accurate, data-efficient and achieving the best overall performance for the goal reaching success rate and safety.\footnote{The code is available at \url{https://github.com/mengyuest/density_planner}}

\section{Related work}
Reachability analysis has been a powerful tool for system verification. The related literature has been extensively studied in~\cite{chen2018hamilton,liu2019algorithms,arch,agha2018survey}. However, few of those have been tackling the problem of calculating reachable set density distribution. Hamilton Jacobian PDE has been used to derive the exact reachable sets in~\cite{mitchell2005time,chen2018hamilton,bansal2020deepreach}, but this approach does not compute the density. Many data-driven methods can compute reachable sets with probabilistic guarantees using scenario optimization~\cite{devonport2020estimating,xue2020pac}, convex shapes~\cite{liebenwein2018sampling,lew2020sampling,berndt2021data}, support vector machines~\cite{allen2014machine,rasmussen2017approximation}, and nonparametric methods~\cite{devonport2020data,thorpe2020data}. However, they cannot estimate state density distribution. \cite{fridovich2020confidence} estimates human policy distribution using a probabilistic model but requires state discretization. \cite{majumdar2014convex} uses the Liouville equation to maximize the backward reachable set for only polynomial system dynamics. In~\cite{abate2007probabilistic}, the authors discretize the system to Markov Chains (MC) and perform probabilistic analysis on MC. This approach is computation-heavy for online safety checks.

Recently, with Neural Networks (NN) development, there has been a growing interest in studying worst-case reachability for NN~\cite{katz2017reluplex,katz2019marabou,xiang2018output,yang2020reachability,vincent2020reachable,meng2021learning} or NN-controlled systems~\cite{ivanov2019verisig,tran2020nnv,fan2020reachnn,hu2020reach,everett2021efficient}. Among those, \cite{meng2021learning} leverages the exact reachability method~\cite{vincent2020reachable} and the Liouville Theorem to perform reachability analysis and reachable set density distribution. This approach finds the probability density function transport equation by solving the Liouville PDE using NN. It shows high accuracy in density estimation compared with histogram, kernel density estimation~\cite{chen2020optimal}, and Sigmoidal Gaussian Cox Processes methods~\cite{donner2018efficient}. Hence, we choose this approach to verify the autonomous systems' safety and to conduct safe motion planning.

There have been various motion planning techniques for autonomous systems, and we refer the interested readers to these surveys~\cite{gonzalez2015review,paden2016survey,goerzen2010survey}. Most approaches use sampling-based algorithms~\cite{karaman2011sampling}, state lattice planners~\cite{pivtoraiko2009differentially,mcnaughton2011motion}, continuous optimization~\cite{qian2016optimal,chen2018foad}, and deep neural networks~\cite{chen2018parallel,zeng2019end}. Reachable sets have also been used for safe motion planning for autonomous systems~\cite{shkolnik2009reachability,gerdts2011avoidance,kousik2020bridging,manzinger2020using}. However, worst-case reachability-based methods only treat reachability as a binary ``yes" or ``no" problem without considering the density distribution of the reachable states. This boolean reachability setting makes the reachability-based motion planner conservative when the collision is inevitable in the worst case (but only happens at a very low probability) thus the system cannot reach the goal state. In this paper, we integrate the density-based reachability estimation method in~\cite{meng2021learning} with model predictive control to improve the goal reaching success rate while enforcing the systems' safety in high probability. 


\section{Problem formulation}
\label{sec:formulation}
Consider a controlled system $\dot{q}=f(q, u)$ where $q\in \mathcal{Q}\subseteq\mathbb{R}^d$ denotes the system state (e.g., position and heading) and $u\in\mathcal{U}\subseteq\mathbb{R}^z$ denotes the control inputs (e.g., thrust and angular velocity). For a given control policy $\pi: \mathcal{Q}\to\mathcal{U}$, the system becomes an autonomous system $\dot{q}=f(q,\pi(q))=f_\pi(q)$ that the future state $q_t$ at time $t$ will only depends on the initial state $q_0$. We assume the initial state $q_0\in\mathcal{Q}_0\subseteq \mathcal{Q}$. Then, the forward reachable set at time $t$ is defined as: 
\begin{equation}
    \mathcal{Q}_t = \left\{q_t \ | \ q_0\in \mathcal{Q}_0,\, \dot{q}=f_\pi(q)\right\}
    \label{eq:forward-set}
\end{equation}

Assume the initial state $q_0$ follows a distribution $\mathcal{D}$ with the support $\mathcal{Q}_0$. Given obstacles $\{\mathcal{O}_i\subseteq \mathbb{R}^p\}_{i=1}^{M}$ in the environment, we aim to compute the probability for colliding with obstacles and the forward probabilistic reachability defined below: 
\begin{definition}[Collision probability estimation]
Given a system $\dot{q}=f_\pi(q)$ with initial state distribution $\mathcal{D}$, compute the probability for states colliding with an obstacle $\mathcal{O}$ at time $t$: $\text{P}_t(\mathcal{O})=\text{Prob}\{q_0\sim\mathcal{D},\,\dot{q}=f_\pi(q),\,\Pi(q_t)\in\mathcal{O}\}$ where $\Pi:\mathcal{Q}\to\mathbb{R}^p$ projects the system state to the space that the obstacle $\mathcal{O}$ resides.
\label{def:1}
\end{definition}

\begin{definition}[Forward probabilistic reachability estimation]
Given a system $\dot{q}=f_\pi(q)$ with initial state distribution $\mathcal{D}$, for each time step $t$, estimate the forward reachable set $\mathcal{Q}_t$ and the probability distribution $\{(\mathcal{A}_i,P_t(\mathcal{A}_i))\}_{i=1}^{N_t}$. Here $\mathcal{A}_1,\,...,\,\mathcal{A}_{N_{t}}$ is a non-overlapping partition for $\mathcal{Q}_t$, i.e., $\mathcal{A}_i \cap \mathcal{A}_j = \varnothing, \forall i\neq j$,  $\mathop{\cup}\limits_{i=1}^{N_{t}} \mathcal{A}_i=\mathcal{Q}_t$. 
\label{def:2}
\end{definition}

Assume $\pi$ is a tracking controller: $\pi(q) = u(q, q^{ref})$ with a reference trajectory $\{q^{ref}_t\}_{t=1}^T$ of length $T$ first generated from a high-level planner with commands $U^{ref}=\{u_t^{ref}\}_{t=1}^T$. Define the \textit{total collision risk}:
\begin{equation}
\text{Prob(colliding)}=P_c(U^{ref})=\sum\limits_{t=1}^T\sum\limits_{i=1}^M P_t(\mathcal{O}_i)
\label{eq:prob-coll}
\end{equation}
We are interested in the following problem: 

\begin{definition}[Safety verification and planning problem]
Given a system $\dot{q}=f_\pi(q)$ with initial state distribution $\mathcal{D}$ and reference control commands $U^{ref}$, verify the total collision risk $P_c(U^{ref})\leq \gamma$, where $\gamma$ is a tolerant collision risk threshold. If not, plan a new command $\tilde{U}^{ref}$ to ensure $P_c(\tilde{U}^{ref})\leq \gamma$  
\label{def:3}
\end{definition}

In this paper, the details about the dynamic systems $\dot{q}=f_\pi(q)$ and controllers $\pi(q) = u(q, q^{ref})$ are listed in the Appendix~\ref{appendix:a} ~\ref{appendix:b}.

\section{Technical approaches}

Inspired by \cite{meng2021learning}, we design a sample-based approach to compute the reachability and the density distribution for the system described in the previous section and further leverage these results for trajectory planning for autonomous systems.

\subsection{Data-driven reachability and density estimation}
\label{sec:sampled-approach}

Our framework is built on top of a recently published density-based reachability analysis method~\cite{meng2021learning}. From the collected trajectory data, \cite{meng2021learning} learns the system flow map and the state density concentration function jointly, guided by the fact that the state density evolution follows the Liouville partial differential equation (PDE). With the set-based reachability analysis tools RPM~\cite{vincent2020reachable}, they can estimate the bound for the reachable set probability distribution.\footnote{For details about computing the probability of the reachable state, we refer the interested readers to~\cite{meng2021learning}(Appendix B).}

For the autonomous system defined in Sec.~\ref{sec:formulation}, we denote the density function $\rho:\mathcal{Q}\times \mathbb{R}\to \mathbb{R}^{\geq 0}$ which measures how states distribute in the state space at a specific time step. The density function is completely determined by the underlying dynamics $f_\pi$ and the initial density map $\rho_0:\mathcal{Q}\to \mathbb{R}^{\geq 0}$ according to the Liouville PDE~\cite{ehrendorfer1994liouville}. 
\begin{equation}
    \frac{\partial \rho}{\partial t} + \nabla \cdot (\rho \cdot f_\pi) = 0, \quad \rho(q, 0) = \rho_0(q)
    \label{eq:liouville}
\end{equation}

We define the flow map $\Phi:\mathcal{Q}\times \mathbb{R}\to \mathcal{Q}$ such that $\Phi(q_0,t)$ is the state at time $t$ starting from $q_0$ at time 0. The density along the trajectory $\Phi(q_0, t)$ is an univariate function of $t$, i.e., $\rho(t)=\rho(\Phi(q_0,t),t)$. If we consider the augmented system with states $[q, \rho]$, from Eq.~\ref{eq:liouville} we can get the dynamics of the augmented system:
\begin{equation}
\begin{bmatrix}
    \dot{q} \\ \dot{\rho}
\end{bmatrix} = 
\begin{bmatrix}
    f_\pi(q) \\ -\nabla\cdot f_\pi(q){\rho}
\end{bmatrix}
    \label{eq:liouode}
\end{equation}

To compute the state and the density at time $T$ from the initial condition $[q_0, \rho_0(q_0)]$, one can solve the Eq.~\ref{eq:liouode} and the solution at time $T$ will give the desired density value. To accelerate the computation process for a large number of initial points, we use neural networks to estimate the density $\rho(q,t)$ and the flow map $\Phi(q,t)$. Details for the network training are introduced in Sec. 3 of~\cite{meng2021learning}.

\subsection{Reach set probability estimation}
\label{sec:prob-approach}
As mentioned in~\cite{meng2021learning}, when the system state is high ($\geq 4$), it is either infeasible (due to the numerical issue in computing for polyhedra) or too time-consuming to generate RPM results for probability estimation. The state dimension will become 7 $\sim$ 10 for a 2D car control or a 3D hovercraft control problem after including the reference control inputs. If we use other worst-case reachability analysis tools such as~\cite{everett2021efficient} to compute the probability, the reachable set will be too conservative 
and the planner will not return a feasible solution (other than stop) because the reachable states will occupy the whole state space regardless of the choice of the reference controls. Therefore, we use a sample-based approach to estimate the probability of the reachable sets, as introduced in the following.


To estimate the probability in Prob.~\ref{def:1}, we first uniformly sample initial states $\{q_0^i\}_{i=1}^{N_s}$ from the support of the distribution $\rho_0$ and use the method in Sec.~\ref{sec:sampled-approach} to estimate the future states and the corresponding densities at time $t$ denoted as $\{(q^i_t,\rho_t(q^i_t))\}_{i=1}^{N_s}$. We approximate the forward reachable set $\mathcal{Q}_t$ defined in Eq.~\ref{eq:forward-set} as the convex hull of $\{q_t^i\}_{i=1}^{N_s}$, and denote it as $\mathcal{CH}_t$. Then, based on $\{(q^i_t,\rho_t(q^i_t))\}_{i=1}^{N_s}$, we use the linear interpolation to estimate the density distribution $\hat{\rho}_{t}(\cdot)$ at time $t$. Finally, we uniformly sample points $q_{s}$ within the convex hull $\mathcal{CH}_t$, and the probability for the system reaching $\mathcal{A}$ can be computed as: 
\begin{equation}
    \text{Prob}(q_t\in\mathcal{A}) \approx \frac{\sum_{q_s\sim \mathcal{CH}_t}\mathbbm{1}\{q_s\in \mathcal{A}\} \hat{\rho}_t(q_s)}{\sum_{q_s\sim\mathcal{CH}_t}\hat{\rho}_t(q_s)}
\label{eq:interp}
\end{equation}

Here are some remarks for our approach. The probabilistic guarantee about estimation accuracy is provided in~\cite{meng2021learning}[Appendix A]. Besides, our approach will return probability zero if the ground truth probability of reaching $\mathcal{A}$ is zero. Moreover, compared to the set-based approach RPM, which has poor scalability because of the number of polyhedral cells growing exponentially to the system state dimension, our sample-based approach is fast, and the runtime can be controlled by selecting different numbers of sampled points as a trade-off between efficiency and accuracy.  

\subsection{Motion planning based on reachability analysis}
\label{sec:plan-approach}

After estimating the reachable states and density for the autonomous system under a reference trajectory, we utilize the results to plan feasible trajectories to ensure the collision probability is under a tolerable threshold.

In this paper, the reference trajectory is generated using nonlinear programming (NLP). Given the origin state $q_{origin}$, the destination state $q_{dest}$, the physical constraints $u_{min} \leq u \leq u_{max}$ and $M$ obstacles $\{(x^o_i, y^o_i)\}_{i=0}^{M-1}$ (with radius $\{r_i\}_{i=0}^{M-1}$) in the environment, discrete time duration $\Delta_t$ and total number of timesteps $T$, we solve an NLP using CasADI~\cite{andersson2019casadi} to generate a reference trajectory, which consists of $N$ trajectory segments $\xi_{0},...,\xi_{N-1}$ (each segment $\xi_j$ has length $L$ and is generated by $q_{j\cdot L}$ and $u_j$). The details about this nonlinear optimization formulation can be found in Appendix~\ref{appendix:c}

Then for each segment $\xi_j$, with the uncertainty and disturbances considered, we use the approach in Sec.~\ref{sec:prob-approach} to estimate the system's reachable states as well as their density. If the total collision risk defined in Eq.~\ref{eq:prob-coll} is below a predefined threshold ($10^{-4}$ in our case), we call the current trajectory ``safe". Otherwise, we call the trajectory ``unsafe" and adjust for the current trajectory segment. Notice that the traditional reachability-based planning is just a special case when we set this threshold to 0.

To ensure fast computation for the planning, we use the perturbation method to sample candidate trajectory segments around this ``unsafe" trajectory segment $\xi_j$ (by adding $\Delta u$ to the reference control commands) and again use the method in Sec.~\ref{sec:prob-approach} to verify whether the candidate is ``unsafe", until we find one segment $\tilde{\xi}_j$ that is ``safe", and then we conduct the NLP starting from the endpoint of the segment $\tilde{\xi}_{j}$. We repeat this process until all the trajectory segments are validated to be ``safe". The whole process is summarized in Algorithm~\ref{algor:planning}.

Given enough sampled points with guaranteed correctness in approximating state density and forward reachable set, the algorithm is sound because the produced control inputs will always ensure the system is ``safe". However, our algorithm is not complete because: (1) in general, the nonlinear programming is not always feasible, and (2) the perturbation method might not be able to find a feasible solution around the ``unsafe" trajectory. The first point can be addressed by introducing slack variables to relax for the safety and goal-reaching constraints. The second point can be tackled by increasing the tolerance probability threshold of collision.

\begin{algorithm}
	\caption{Reachability-based Planning Algorithm}
	\begin{flushleft}
	\textbf{Input:} Origin $S_0$, destination $S_N$, NLP constraints \\
	\textbf{Output:} Reference trajectories $\xi_0, \xi_1, \cdots, \xi_{N-1}$
	\end{flushleft}
	\begin{algorithmic}[1]
	    \State $i\leftarrow 0$
		\While {$i<N$}
    		\State Generate segments $\xi_{i}, ..., \xi_{N-1}$ from $S_i$ to $S_N$ using NLP
    		\For {$j=i:N$}
        		\State Use the method in Sec.~\ref{sec:prob-approach} to check whether the trajectory segment $\xi_j$ is ``safe".
        		\If {The trajectory is ``safe"}
        		    \State Continue
        		\Else
        		    \State Perturb the segment $\xi_j$ to search for a possible ``safe" segment $\tilde{\xi}_j$ (goes from $S_j$ to close to $S_{j+1}$).
        		    \State $\xi_j \leftarrow \tilde{\xi}_j$
        		    \State $S_{j+1}\leftarrow \tilde{S}_{j+1}$
        		\EndIf
    		\EndFor
            \Comment{By far, $S_0\to S_{j+1}$ is ``safe"}
    		\State $i \leftarrow j+1$ \Comment{Next step will inspect $S_{j+1}\rightarrow \cdots S_{N}$}
		\EndWhile
	\end{algorithmic}
	\label{algor:planning}
\end{algorithm}


\section{Experiments}
We investigate our approach in autonomous driving and hovercraft navigation applications under the following setup: given an environment with an origin point, a destination region, and obstacles, the goal for the agent at the origin point is to reach the destination while avoiding all the obstacles. Notice that this is a very general setup to encode the real-world driving scenarios because: (1) the road boundaries and other irregular-shaped obstacles can be represented by using a set of obstacles, and (2) other road participants (pedestrians, other driving cars) can be modeled as moving obstacles. Here we consider only the center of mass of the car/hovercraft in rendering reachable sets and planning (we can bloat the radius of the obstacle to take the car/hovercraft length and width into account). In Sec.~\ref{sec:case-1}, we evaluate the reachability and density for the system under a fixed reference trajectory. In Sec.~\ref{sec:case-2}, we leverage the reachability and density result to do trajectory re-planning when the system is ``unsafe".

We collect 50,000 trajectories from the simulator, with randomly sampled initial states, reference trajectories, and disturbances. Each trajectory has 50 timesteps with a duration of 0.02s at each time step. Then, we select 40,000 for the training set and 10,000 for the evaluation set and train a neural network for estimating the future states and the density evolution mentioned in~\cite{meng2021learning}. We use a fully connected ReLU-based neural network with 3 hidden layers and 128 hidden units in each layer. We train the neural network for 500k epochs, using stochastic gradient descent with a batch size of 256 and a learning rate of 0.1. The code is implemented in PyTorch~\cite{paszke2019pytorch}, and the training takes less than an hour on an NVidia RTX 2080Ti GPU.

\subsection{Reachable states and density estimation} 
\label{sec:case-1}
In this section, we first conceptually show how our approach of estimating reachable states and density can benefit safety-critical applications. As depicted in Fig.~\ref{fig:demo}, a car plans to move to the destination (the red arrow) while avoiding all the obstacles on the road. The initial state of the car (X,Y position, and heading angle) and the disturbance follow a Gaussian distribution. The high-level motion planner has already generated a reference trajectory (the blue line in Fig.~\ref{fig:demo}), with the uncertainty owing to the initial state estimation error and the disturbance. We will show that our approach can estimate the state density distribution and reachable state accurately and can help to certify that the planned reference trajectory is not colliding with obstacles in high probability.

\subsubsection{Visualizations of the estimated reachability and density heatmap}
\label{sec:vis_heatmap}
\begin{figure}[!htbp]
\begin{subfigure}[b]{0.230\textwidth}
\includegraphics[width=1.0\textwidth]{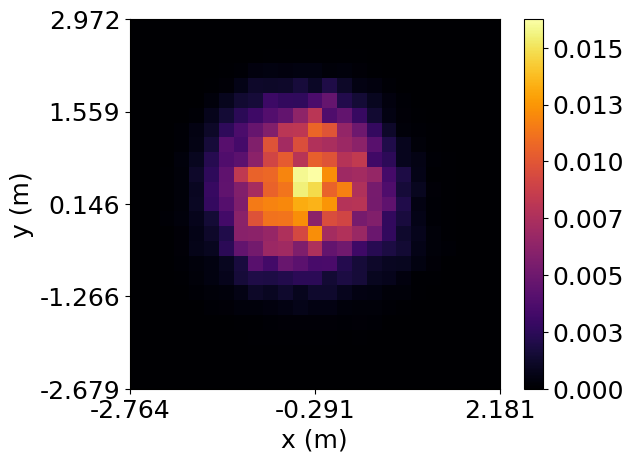} \hfill
\caption{t=0.0s}
\end{subfigure}
\begin{subfigure}[b]{0.230\textwidth}
\includegraphics[width=1.0\textwidth]{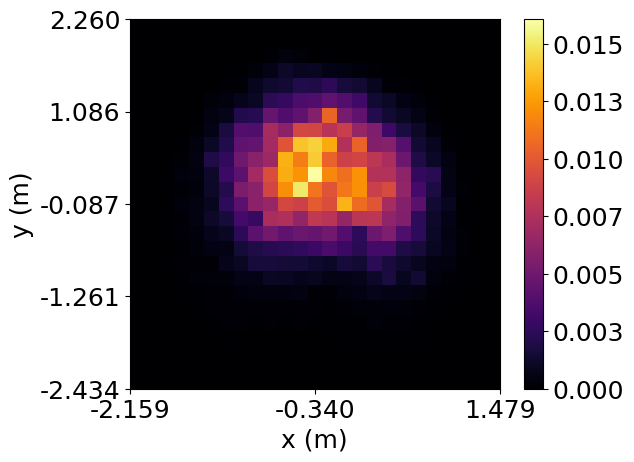} \hfill
\caption{t=0.3s}
\end{subfigure}
\begin{subfigure}[b]{0.230\textwidth}
\includegraphics[width=1.0\textwidth]{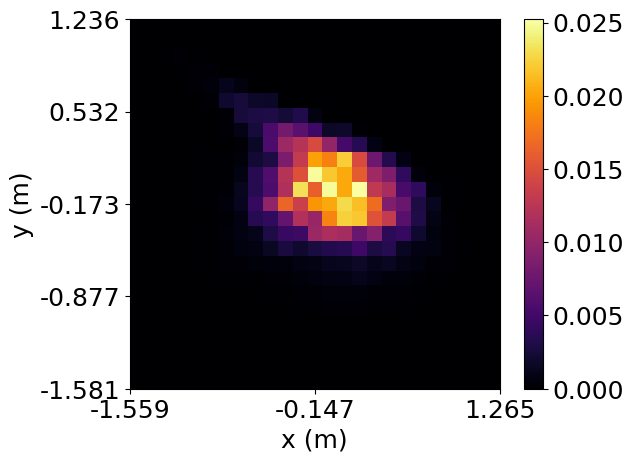} \hfill
\caption{t=0.8s}
\end{subfigure}
\begin{subfigure}[b]{0.230\textwidth}
\includegraphics[width=1.0\textwidth]{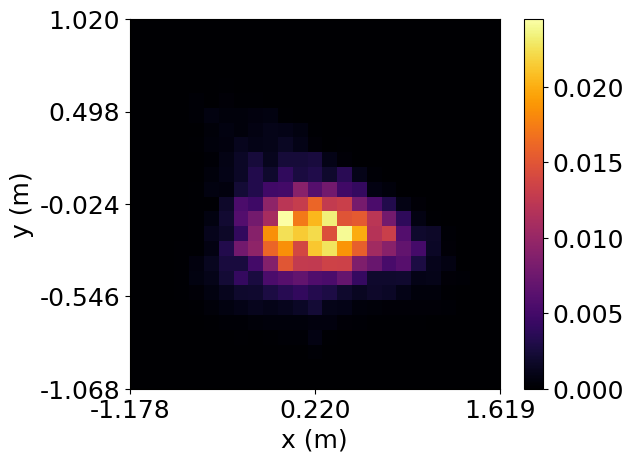} \hfill
\caption{t=1.0s}
\end{subfigure}
\begin{subfigure}[b]{0.51\textwidth}
\includegraphics[width=1.0\textwidth]{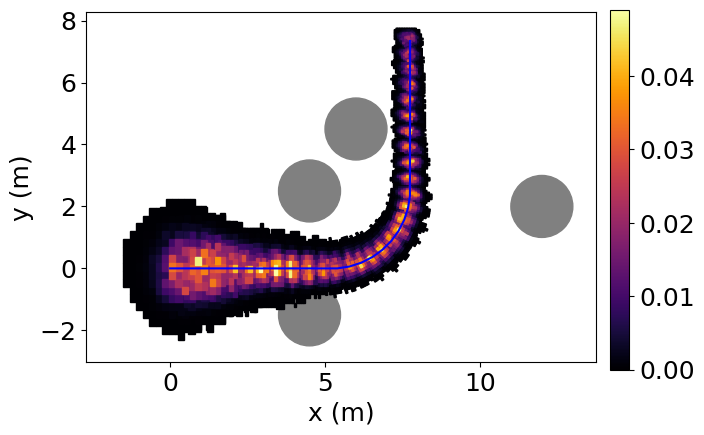} 
 \hfill
\caption{Density distribution on trajectories.}
\end{subfigure}
\begin{subfigure}[b]{0.44\textwidth}
\includegraphics[width=1.0\textwidth]{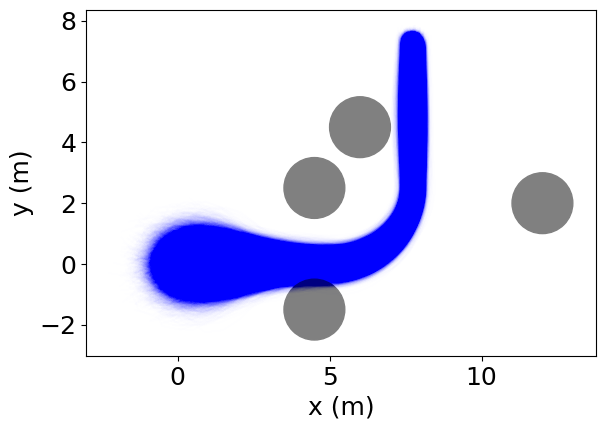} 
 \hfill
\caption{Simulated trajectories.}
\end{subfigure}
\centering
\caption{Estimated density (for states in (a)-(d) and along the trajectory in (e)) for the car model. The states are shown to concentrate on reference trajectory (blue line in (e)), and the collision risk is very low.}
\label{fig:heatraj}
\end{figure}

Using the method introduced in Sec.~\ref{sec:prob-approach}, we can first estimate the tracking error density distribution and marginalize it to the 2D XY-plane to get the probability heatmaps (as shown in Fig.~\ref{fig:heatraj}(a)-(d)). Then we can transform it to the reference trajectory and check whether it has an intersection with the obstacles in the environment (as shown in Fig~\ref{fig:heatraj}(e)). 

\begin{figure}[!htbp]
\begin{subfigure}[b]{0.230\textwidth}
\includegraphics[width=1.0\textwidth]{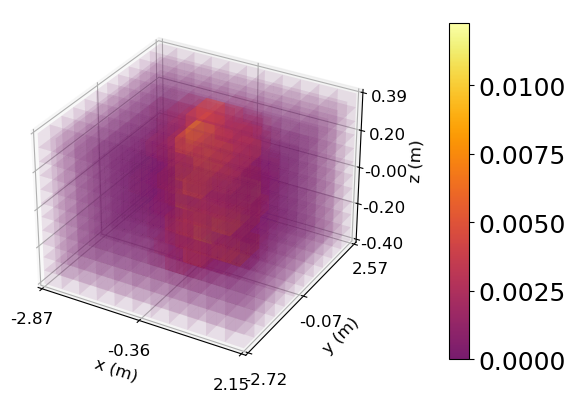} \hfill
\caption{t=0.1s}
\end{subfigure}
\begin{subfigure}[b]{0.230\textwidth}
\includegraphics[width=1.0\textwidth]{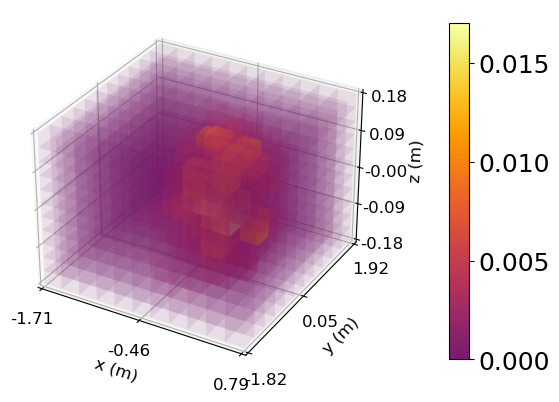} \hfill
\caption{t=0.3s}
\end{subfigure}
\begin{subfigure}[b]{0.230\textwidth}
\includegraphics[width=1.0\textwidth]{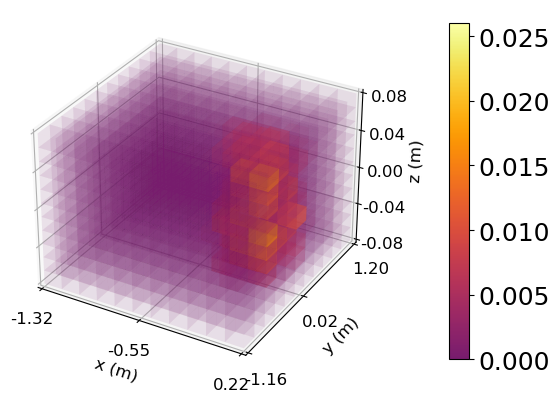} \hfill
\caption{t=0.8s}
\end{subfigure}
\begin{subfigure}[b]{0.230\textwidth}
\includegraphics[width=1.0\textwidth]{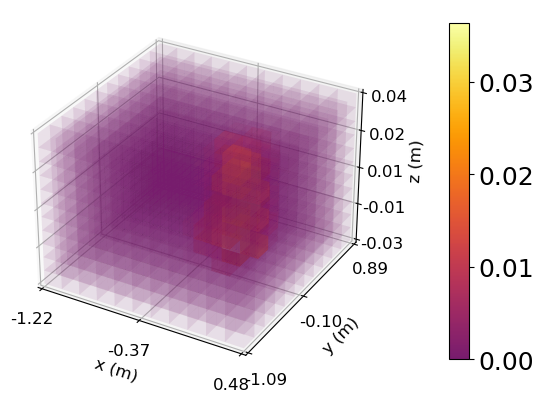} \hfill
\caption{t=1.0s}
\end{subfigure}
\begin{subfigure}[b]{0.48\textwidth} 
\includegraphics[width=1.0\textwidth]{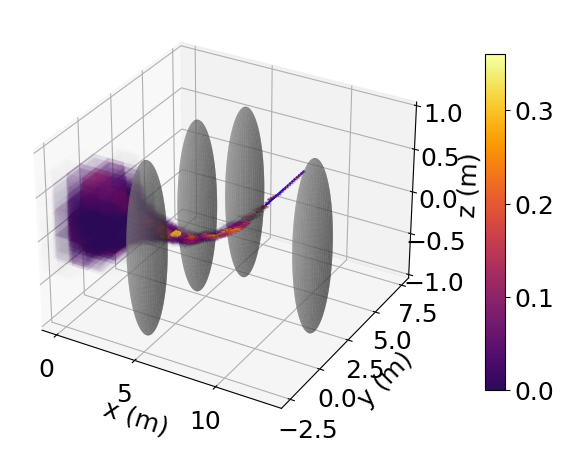}
 \hfill
\caption{Density distribution on trajectories.}
\end{subfigure}
\begin{subfigure}[b]{0.42\textwidth} 
\includegraphics[width=1.0\textwidth]{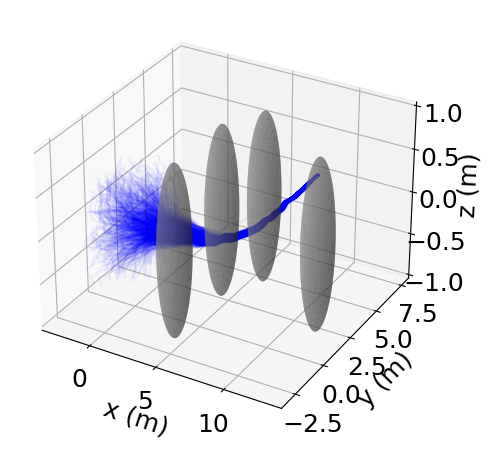}
 \hfill
\caption{Simulated trajectories.}
\end{subfigure}
\centering
\caption{Estimated density (for states in (a)-(d) and along the trajectory in (e)) for the hovercraft model. The states are shown to concentrate on reference trajectory (blue line in (e)), and the collision probability is very low.}
\label{fig:heatraj3d}
\end{figure}

To verify the correctness of our estimated reachable states and density, we also sample a large number of states from the initial state distribution and use the ODE to simulate actual car trajectories, as shown in Fig.~\ref{fig:heatraj}(b). Comparing Fig.~\ref{fig:heatraj}(a) with Fig.~\ref{fig:heatraj}(b), we find out in both cases that the vehicle will have a collision with the bottom obstacle. In addition, our density result also shows that the risk of the collision is very low ($\text{Prob(colliding)}\leq 10^{-4}$ as shown in Sec.~\ref{sec:case-1-dens}), which is reasonable because the majority of the states will be converging to the reference trajectory (as indicated from Fig.~\ref{fig:heatraj}(a)-(d)). Only a few outlier trajectories will intersect with the obstacle. We also conduct this experiment with the Hovercraft system (3D scenarios), where the results in Fig.~\ref{fig:heatraj3d} reflect similar contraction behaviors, and the probability of colliding with the obstacles is very low (thus, we do not need to do planning in this stage).

Visually, our results are more informative than the pure reachability analysis because ours reflects the tracking controller's contraction behavior and illustrates that the colliding event is in very low probability. The following subsection will further quantify this probability and compare it with a traditional probability estimation method.

\subsubsection{Comparison with Monte-Carlo based probability estimation}
\label{sec:case-1-dens}

\begin{figure}
\centering
\begin{minipage}{.45\textwidth}
  \centering
  \includegraphics[width=.95\linewidth]{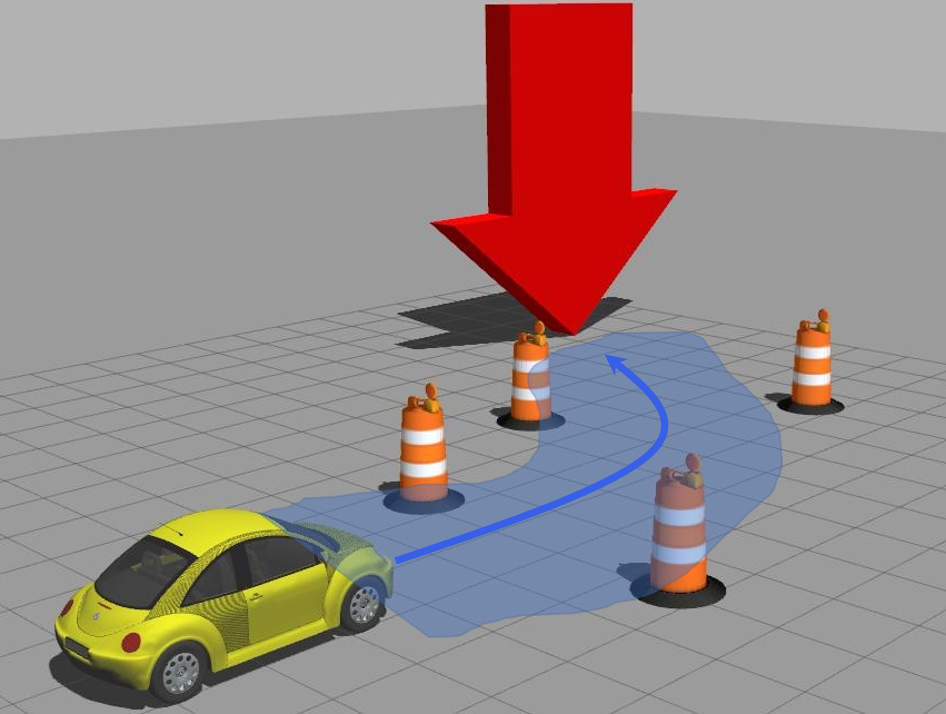}
  \captionof{figure}{A vehicle plans to reach the red arrow while avoiding all the obstacles. Blue region shows the reference trajectory and uncertainty.}
  \label{fig:demo}
\end{minipage}%
\hfill
\begin{minipage}{.53\textwidth}
  \centering
  \includegraphics[width=.95\linewidth]{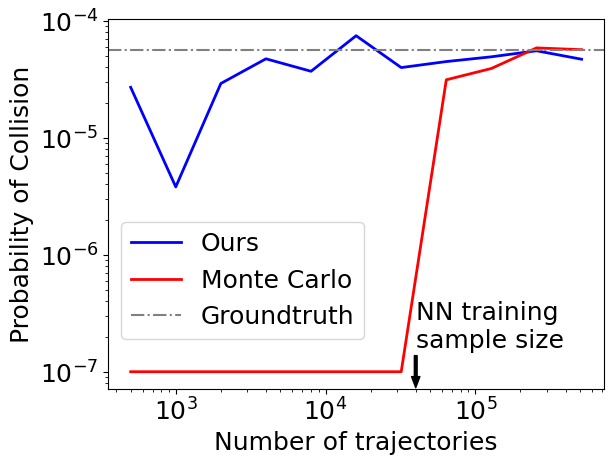}
  \captionof{figure}{Estimation of collision probability with respect to sample size.}
  \label{fig:prob_est}
\end{minipage}
\end{figure}

Our visualization result in Sec.~\ref{sec:vis_heatmap} reflects that under some initial conditions, the vehicle might hit the obstacle. Although Fig.~\ref{fig:heatraj} shows that the likelihood of the clash is very low, we want to quantify the risk of the collision to benefit future decisions (e.g., choosing the policy with the lowest collision probability or with the lowest value at risk (VaR) ~\cite{majumdar2020should}). However, it is intractable to derive the ground-truth probability of collision for general non-linear systems. Therefore, we compare our estimation result with the Monte Carlo approximation (this is done by generating a considerable amount of simulations and counts for the frequency of the collision.

We try different numbers of samples (from 500 to 512000) and compare our approach to the Monte Carlo estimation. As shown in Fig.~\ref{fig:prob_est}, the groundtruth probability of collision (where we approximated by sampling $5\times 10^7$ trajectories and compute the collision rate) is approximately $5\times 10^{-5}$. The Monte Carlo approach fails to predict meaningful probability results until increasing the sample size to 64000. In contrast, our approach can give a non-trivial probability estimation using only 500 examples, less than 0.01X the samples needed for the Monte Carlo approach. 

The black vertical arrow in Fig.~\ref{fig:prob_est} corresponds to the 40000 sample size (the number of samples we have used offline for our neural network training). The corresponding result is already as stable as the Monte Carlo approach which requires more than 64000 samples. Furthermore, our approach can be adapted to any initial condition for the same car dynamic system without retraining or fine-tuning, making it possible for downstream tasks like online planning, as introduced in the following section.

In terms of the computation time (for 512000 points), our approach requires 2.3X amount of time (31.8s) as needed for the Monte Carlo method (14.1s), mainly because that the Delaunay Triangulation method~\cite{lee1980two} used for state-space partition in the linear interpolation has the complexity of $O(N^{\lfloor d/2 \rfloor})$~\cite{toth2017handbook} for $N$ data points in the $d$-dimension system. Thus the run time will grow about quadratically to the number of sample points in our case (state dimension=4). With 1000 sampled points (as used in the rest of the experiments), our method takes $\sim 2$ seconds, which is acceptable. One alternative solution to accelerate the computation is to use Nearest Neighbor interpolation for density estimation.

\subsection{ Online planning via reachable set density estimation}
\label{sec:case-2}
When the probability of collision is higher than the threshold ($10^{-4}$ in our experiment setting), we need to use the planning algorithm to ensure the safety of the autonomous system under uncertainty and disturbance. We first show how the planning algorithm works in Sec.~\ref{sec:case-2-demo}, and then quantitative assessment of our algorithm is conducted in Sec.~\ref{sec:eval_plan}.

\subsubsection{Demonstration for an example}
\label{sec:case-2-demo}
\begin{figure}[!htbp]
\begin{subfigure}[b]{0.32\textwidth}
\includegraphics[width=1.0\textwidth]{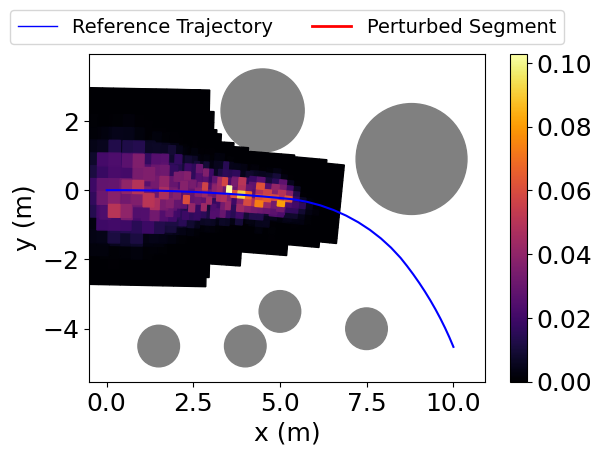} \hfill
\caption{t=0, ``unsafe" detected. Prob(colliding) $\geq 10^{-4}$}
\end{subfigure}
\begin{subfigure}[b]{0.32\textwidth}
\includegraphics[width=1.0\textwidth]{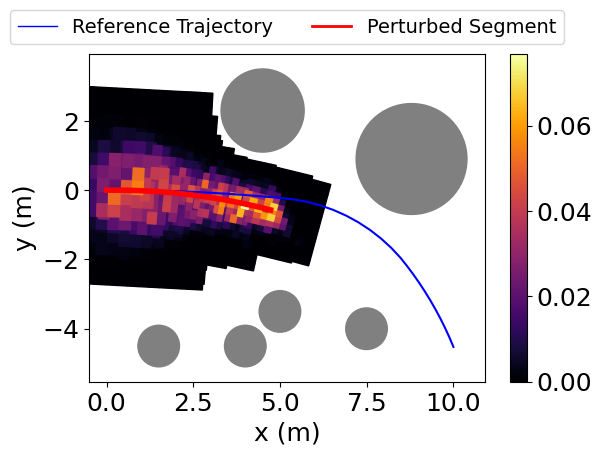} \hfill
\caption{t=0, perturb to ``safe". Prob(colliding) $\leq 10^{-4}$}
\end{subfigure}
\begin{subfigure}[b]{0.32\textwidth}
\includegraphics[width=1.0\textwidth]{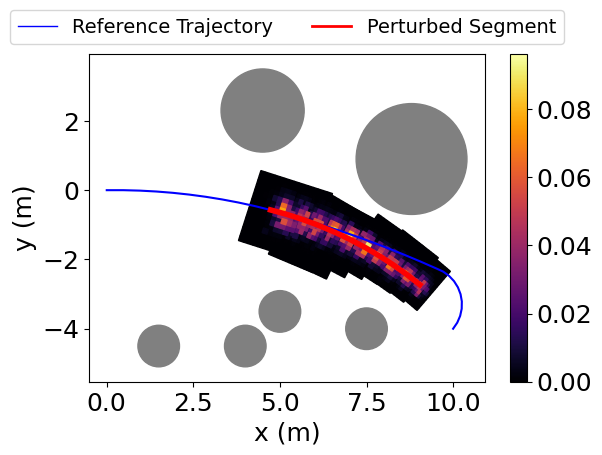} \hfill
\caption{t=1, perturb to ``safe". Prob(colliding) $\leq 10^{-4}$}
\end{subfigure}
\centering
\caption{Demonstration for the re-planning algorithm for the 2D car experiment. For each trajectory segment, if the probability of collision is higher than a predefined threshold, we denote the trajectory segment as ``unsafe", otherwise, we denote the trajectory segment as ``safe".}
\label{fig:demo_planning}
\end{figure}

\begin{figure}[!htbp]
\begin{subfigure}[b]{0.32\textwidth}
\includegraphics[width=1.0\textwidth]{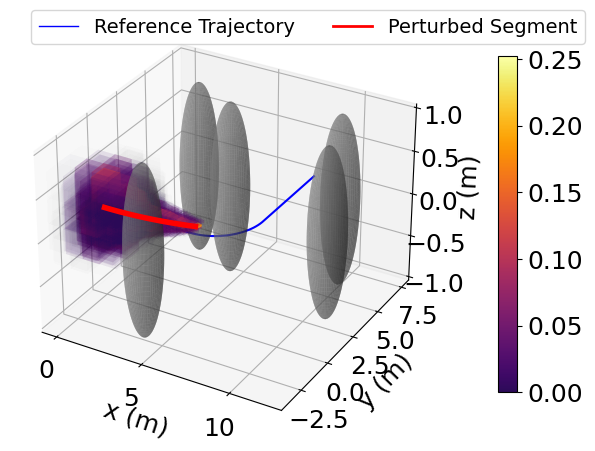} \hfill
\caption{t=0, perturb to ``safe". Prob(colliding) $\leq 10^{-4}$}
\end{subfigure}
\begin{subfigure}[b]{0.32\textwidth}
\includegraphics[width=1.0\textwidth]{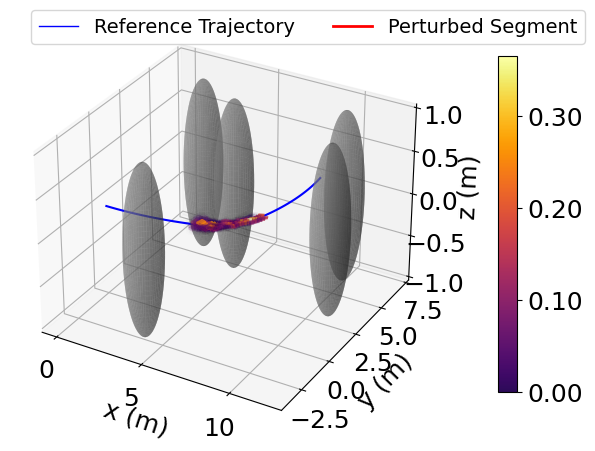} \hfill
\caption{t=1, ``safe" is verified. Prob(colliding) $\leq 10^{-4}$}
\end{subfigure}
\begin{subfigure}[b]{0.32\textwidth}
\includegraphics[width=1.0\textwidth]{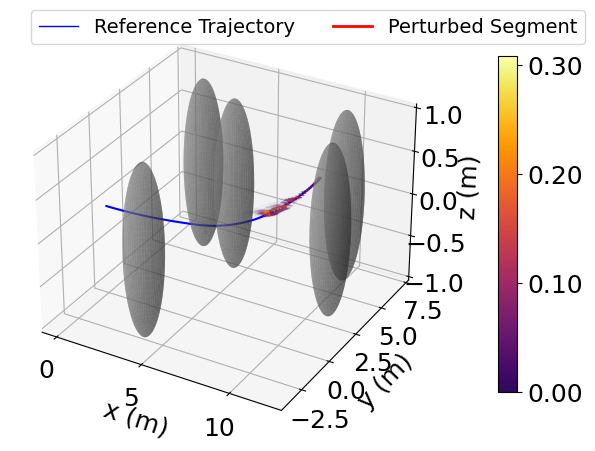} \hfill
\caption{t=2, ``safe" is verified. Prob(colliding) $\leq 10^{-4}$}
\end{subfigure}
\centering
\caption{Demonstration for the re-planning algorithm for the 3D hovercraft experiment. For each trajectory segment, if the probability of collision is higher than a predefined threshold, we denote the trajectory segment as ``unsafe", otherwise, we denote the trajectory segment as ``safe".}
\label{fig:demo_plannin_hov}
\end{figure}

We conduct experiments to demonstrate how our proposed reachability-based planning framework works for autonomous driving cars and hovercraft applications. In Fig.~\ref{fig:demo_planning}, a car is moving from left to right of the map while avoiding collisions with obstacles. After checking for the first segment's safety (Fig.~\ref{fig:demo_planning}(a)), the algorithm finds out the probability of the collision is higher than $10^{-4}$. Thus it starts to plan for the first segment (the red line in Fig.~\ref{fig:demo_planning}(b)) around the reference trajectory (the blue line in Fig.~\ref{fig:demo_planning}(b)). Moreover, after it updates the reference trajectory using the NLP solver (the blue line in Fig.~\ref{fig:demo_planning}(c)) based on the perturbed segment, the algorithm detects the next segment as ``unsafe." Hence it plans again to enforce the collision probability is below $10^{-4}$ (Fig.~\ref{fig:demo_planning}(c)). Another example in a 3D scenario is shown in Fig.~\ref{fig:demo_plannin_hov}, where the hovercraft perturbs for the first segment and then later verifies that the next two segments are all ``safe", hence the whole trajectory in Fig.~\ref{fig:demo_plannin_hov}(c) is ``safe."

\subsubsection{Evaluation of trajectory planning performance}
\label{sec:eval_plan}

To illustrate the advantage of our approach in enforcing the system safety, we design 20 testing environments with randomly placed obstacles (while ensuring the initial reference trajectory is feasible from the NLP solver) for the autonomous car system and the hovercraft control system. We compare our framework with several baseline methods: the “original” approach just uses the initial reference trajectory (without any planning process), the ``$d=?$” approach denotes the distance-based planning methods with the safety distance threshold set as $d$ (the larger $d$ is, the more conservative and safer the algorithm will be, at the cost of infeasible solutions), the ``reach" approach uses estimated reachable tube computed from the sampled convex hull $\mathcal{CH}$ introduced in Sec.~\ref{sec:prob-approach} to do safe planning.

We measure the performances of different methods using two metrics: feasibility and safety. \textit{Feasibility} is defined as the frequency that the algorithm can return a plausible solution (but might be unsafe) to reach the designed destination. Over all feasible solutions, \textit{safety} is defined as the expected collision probability. Intuitively, the feasibility measures the successful rate of the goal reaching, and the ``safety" measures how ``reliable" the planner is.

\begin{figure}[!htbp]
\begin{subfigure}[b]{0.490\textwidth}
\includegraphics[width=1.0\textwidth]{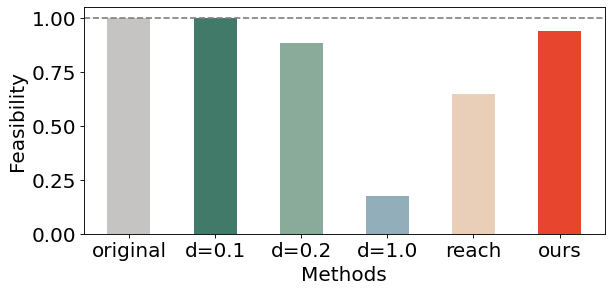} \hfill
\caption{Car model, feasibility comparisons}
\end{subfigure}
\begin{subfigure}[b]{0.490\textwidth}
\includegraphics[width=1.0\textwidth]{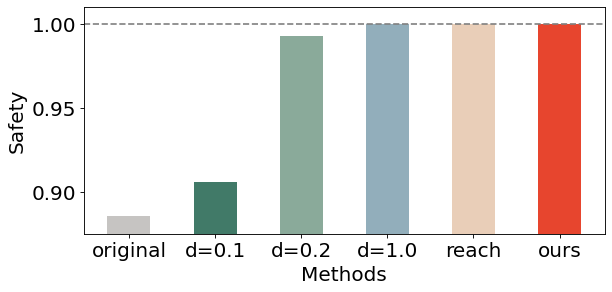} \hfill
\caption{Car model, safety comparisons}
\end{subfigure}\\
\begin{subfigure}[b]{0.490\textwidth}
\includegraphics[width=1.0\textwidth]{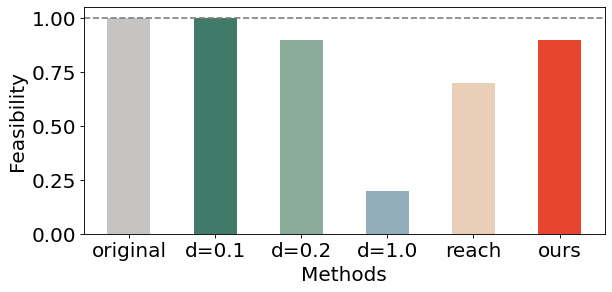} \hfill
\caption{Hovercraft, feasibility comparisons}
\end{subfigure}
\begin{subfigure}[b]{0.490\textwidth}
\includegraphics[width=1.0\textwidth]{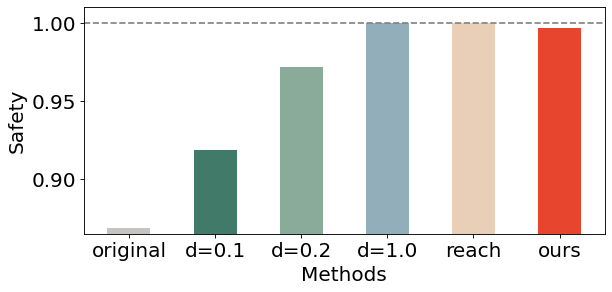} \hfill
\caption{Hovercraft, safety comparisons}
\end{subfigure}\\
\centering
\caption{Feasibility and safety comparisons. Our method achieves the best trade-off between the feasibility and the safety, with close to 100\% safety and 0.2X $\sim$ 4X improvement in feasibility comparing to high-safety methods. Here ``original" uses nonlinear programming for planning
,``$d=?$" denotes the distance-based re-planning with the safety distance $d$, ``reach" uses estimated reachable tube to do re-planning, and ``ours" leverages both the reachable tube and the corresponding density to do re-planning. More details can be found in Sec.~\ref{sec:eval_plan}}
\label{fig:eval_plan}
\end{figure}

As shown in Fig.~\ref{fig:eval_plan}(a)(b) for the car experiments, compared to the ``reach" method (which just uses reachable tubes to do planning) and a distance-based planning method (``$d=1.0$"), our approach can achieve a similar safety rate, while having 0.29 $\sim$ 0.76 higher feasibility (due to less conservative planning). Though we are 0.059 less in feasibility than the ``original" and the ``d=0.1" baselines, they lead to 3772X $\sim$ 4574X collision rates than ours (due to our approach being less aggressive in planning). Compared to the ``$d=0.2$" method, we are 0.06 better in feasibility ($0.94$ vs $0.88$) while being only $0.2X$ in collision rate ($0.000025$ vs $0.000125$), we also get a lower collision rate . A similar trend can also be observed from the hovercraft experiment (Fig.~\ref{fig:eval_plan}(c)(d)). Hence, our approach achieves the best overall performance by considering feasibility and safety.



\subsection{Discussions}
While our approach can accurately estimate the collision probability and the motion planner using our estimated density can achieve the highest goal reaching rate compared to other baselines when enforcing the safety rate above 0.99, we admit there are assumptions and limitations in our method. 

First we assume the neural network can learn a perfect state dynamic and state density evolution, which is not always satisfied due to the model capacity and the complexity of the system. The proof in~\cite{meng2021learning}[Appendix A] shows the generalization error bound for this learning framework, 
which indicates one possible remedy is to collect more training trajectories. 

Besides, we assume the sampled trajectories from the simulator are following the same distribution as for the real world trajectories. This assumption might create biases in the density concentration function and the flow map estimation. One way to resolve this issue is to further fine-tuning the neural network using the real world data at the inference stage.


The first limitation of our approach is lacking guarantee for the convergence of the planning algorithm. The success planning rate of our method depends on the perturbation range (how far the control policy can deviate from the reference policy) and the perturbation resolution (the minimum difference between two candidate policies). There are also optimization-based methods, such as stochastic gradient descent~\cite{amari1993backpropagation}, that can  converge with probabilistic guarantee derived from the Robbins-Siegmund theorem~\cite{robbins1971convergence}. Using optimization-based method for density-based planning is left to our future work. 

The second limitation of our planning framework is the computation time. This is mainly due to our risk computation step in Eq.~\ref{eq:prob-coll}, as mentioned in Sec.~\ref{sec:case-1-dens}. Although our proposed probability computation method can handle higher dimension systems than~\cite{meng2021learning}, the complexity of the Delaunay Triangulation process in our framework grows in the power of $\lfloor d/2\rfloor$ for a $d$-dimensional system. In practice, one can use less number of sample points to reduce the computation time. Another alternative is to use other interpolation methods (e.g., Nearest Neighbor interpolation) for $d$-dimensional space.


\section{Conclusion}
We propose a data-driven framework for probabilistic verification and safe motion planning for autonomous systems. Our approach can accurately estimate collision risk, using only 0.01X training samples compared to the Monte Carlo method. We conduct experiments for autonomous driving and hovercraft control, where the car (hovercraft) with state uncertainty and control input disturbances plans to move to the destination while avoiding all the obstacles. We show that our approach can achieve the highest goal reaching
rate among all approaches that can enforce the safety rate above 0.99. For future works, we manage to develop verification approaches for cases that consider (1) other road participants' presence and intention and (2) more complicated sensory inputs, such as LiDAR measurements or even raw camera inputs.


\section*{Acknowledgement}
The Ford Motor Company provided funds to assist the authors with their research, but this article solely reflects the opinions and conclusions of its authors and not any Ford entity. The authors would also want to thank Kyle Post for providing constructive suggestions regarding experiment designs, figures, and paper writings.

\clearpage
\renewcommand\thesection{\Alph{section}}
\setcounter{section}{0}
\setcounter{figure}{0}

\begin{doublespacing}
{\hspace{-1.6em}\LARGE\textbf{Appendix}}
\end{doublespacing}

\section{Car model dynamic and controller designs}
\label{appendix:a}
The dynamics for the rearwheel kinematic car model \cite{fan2020fast} is:
\begin{equation}
    \dot{q}=\begin{bmatrix}
    \dot{x} \\ \dot{y} \\ \dot{\theta}
    \end{bmatrix} = \begin{bmatrix}
    \cos{\theta} & 0 \\ 
    \sin{\theta} & 0 \\
    0 & 1
    \end{bmatrix} \begin{bmatrix}
    v \\ \omega
    \end{bmatrix}
\end{equation}
where $(x, y)$ denotes the position of the vehicle's center of mass, $\theta$ denotes vehicle's heading angle, and $v, \omega$ are the velocity and angular velocity control inputs. Given a reference trajectory generated from the motion planner $(x^{ref}, y^{ref}, \theta^{ref})^T$, the error for the car model is:
\begin{equation}
\begin{bmatrix}
e_x \\ e_y \\ e_{\theta}
\end{bmatrix} =\begin{bmatrix}
\cos(\theta) & \sin(\theta) & 0 \\
-\sin(\theta) & \cos(\theta) & 0 \\
0 & 0 & 1
\end{bmatrix}
\begin{bmatrix}
x-x^{ref}\\ 
y-y^{ref}\\
\theta-\theta^{ref}
\end{bmatrix}
\end{equation}

The Lyapunov-based controller is designed as:
\begin{equation}
    \begin{cases}
        v = v^{ref} \cos(e_{\theta}) + k_1 e_x + d_{v}\\
        \omega = \omega^{ref} + v^{ref} (k_2 e_y + k_3 \sin (e_{\theta})) + d_{\omega}
    \end{cases}
\end{equation}
where $k_1,k_2,k_3$ are the coefficients for the controller and $d_{v}$ and $d_{\omega}$ are the controller disturbances.

\section{Hovercraft model dynamic and controller designs}
\label{appendix:b}
The hovercraft is the model tested in 3D scenarios. The dynamics for the hovercraft \cite{fan2020fast} is:
\begin{equation}
    \dot{q}=\begin{bmatrix}
    \dot{x} \\ \dot{y} \\ \dot{z} \\ \dot{\theta}
    \end{bmatrix} = \begin{bmatrix}
    \cos{\theta} & 0 & 0 \\ 
    \sin{\theta} & 0 & 0\\
    0 & 1 & 0 \\
    0 & 0 & 1 
    \end{bmatrix} \begin{bmatrix}
    v \\ v_z \\ \omega
    \end{bmatrix}
\end{equation}
where $(x,y,z)$ denotes the 3D position of the hovercraft's center of mass, $\theta$ denotes the heading angle of the hovercraft in the $xy$-plane, $v$ (and $\omega$) denotes the velocity (and angular velocity) in the $xy$-plane, $v_z$ denotes the velocity along the $z$-axis. When a reference trajectory $(x^{ref}, y^{ref}, z^{ref}, \theta^{ref})^T$ is introduced, the error for the car model is: 
\begin{equation}
\begin{bmatrix}
e_x \\ e_y \\ e_z \\ e_{\theta}
\end{bmatrix} =\begin{bmatrix}
\cos(\theta) & \sin(\theta) & 0 & 0 \\
-\sin(\theta) & \cos(\theta) & 0 & 0 \\
0 & 0 & 1 & 0 \\
0 & 0 & 0 & 1 \\
\end{bmatrix}
\begin{bmatrix}
x-x^{ref}\\ 
y-y^{ref}\\
z-z^{ref}\\
\theta-\theta^{ref}
\end{bmatrix}
\end{equation}

The Lyapunov-based controller is designed as:
\begin{equation}
    \begin{cases}
        v = v^{ref} \cos(e_{\theta}) + k_1 e_x + d_{v}\\
        v_z = v_z^{ref} + k_4 e_z + d_{v_z}\\
        \omega = \omega^{ref} + v^{ref} (k_2 e_y + k_3 \sin (e_{\theta})) + d_{\omega}
    \end{cases}
\end{equation}
where $k_1,k_2,k_3,k_4$ are the coefficients for the controller and $d_{v}$, $d_{v_z}$ and $d_{\omega}$ are the corresponding disturbances.

\section{Nonlinear programming for controller synthesize}
\label{appendix:c}
The goal of this section is to find a control sequence $\{u_j\}_{j=0}^{N-1}$ for the car (or the hovercraft, we use ``robot" to represent them in the following context) starting from $q_{origin} \in \mathbb{R}^d$ to reach the goal state $q_{dest} \in \mathbb{R}^d$ in $T$ time steps, while satisfying the physical constraints and avoiding colliding with the surrounding obstacles ($M$ obstacles in total) in the environment. We use the forward Euler method to compute the ODE $\dot{q}=f(q,u)$, with each time step duration as $\Delta t$. Each control input $u_j$ will last for $L=\lceil\frac{T}{N}\rceil$ steps. For the physical constraints, we set up the maximum and minimum allowed value for the control inputs as $u_{max}, u_{min} \in \mathbb{R}^z$. We represent the obstacles as circles (and spheres in 3D scenarios). The $i$-th obstacle has a center position $\bar{q}^o_i \in \mathbb{R}^2$ ($\bar{q}^o_i\in\mathbb{R}^3$ in 3D scenarios) and a radius $r_i \in \mathbb{R}^+$ (we use $\bar{q}_j$ to represent the robot position at time $j$, to distinguish with the full robot state $q_j$). We formulate the optimization process as followed: 
\begin{equation}
    \begin{aligned}
        \min\limits_{u_{0:N-1}} \quad & \sum\limits_{i=0}^{M-1}\sum\limits_{j=0}^{T}\gamma_{i,j}^2 \\
        \text{s.t.} \quad & q_0=q_{origin}\\
        & q_T=q_{dest} \\
        & u_{min} \leq u_j \leq u_{max},\forall j =0...N-1 \\
        & q_{j\cdot L+k+1} = q_{j\cdot L+k} + f(q_{j\cdot L+k}, u_j)\Delta t , \forall j=0...N-1, \forall k=0...L-1\\
        & |\bar{q}_j-\bar{q}^o_i|^2 + \gamma_{i,j}^2 \geq {r_i}^2, \forall i=0...,M-1, \forall j=0,...,T\\
    \end{aligned}
    \label{eq:nlp}
\end{equation}
where the first two constraints make sure the robot starts from the initial point and will reach the goal point, the third and forth constraints enforce the physical constraints and the robot dynamic, and the last constraint ensure the robot will not hit obstacles at any time. For feasibility issues, slack variables $\gamma_{i,j}$ are introduced to relax the collision avoidance constraint (the robot safety will be checked and ensured during the online planning process after this optimization process).

\newpage
\bibliographystyle{plain}
\bibliography{z7_references}

\end{document}